\documentclass[10pt,twocolumn,letterpaper]{article}

   \usepackage[pagenumbers]{iccv} 

\definecolor{iccvblue}{rgb}{0.21,0.49,0.74}
\usepackage[pagebackref,breaklinks,colorlinks,allcolors=iccvblue]{hyperref}

\usepackage{graphicx} 
\usepackage[table]{xcolor} 
\usepackage[accsupp]{axessibility}

\def\paperID{9} 
\def\confName{ICCV}
\def\confYear{2025}

\title{SmolRGPT: Efficient Spatial Reasoning for Warehouse Environments with 600M Parameters}

\author{Abdarahmane Traore, Éric Hervet, Andy Couturier\\
Embia, Computer Science Department, Faculty of Science, Université de Moncton\\
Moncton, NB, Canada\\
{\tt\small eat4651@umoncton.ca}\\
{\tt\small andy.couturier@umoncton.ca}\\
{\tt\small eric.hervet@umoncton.ca}\\
}

\begin{document}
\maketitle
\begin{abstract}
Recent advances in vision-language models (VLMs) have enabled powerful multimodal reasoning, but state-of-the-art approaches typically rely on extremely large models with prohibitive computational and memory requirements. This makes their deployment challenging in resource-constrained environments such as warehouses, robotics, and industrial applications, where both efficiency and robust spatial understanding are critical. In this work, we present SmolRGPT, a compact vision-language architecture that explicitly incorporates region-level spatial reasoning by integrating both RGB and depth cues. SmolRGPT employs a three-stage curriculum that progressively align visual and language features, enables spatial relationship understanding, and adapts to task-specific datasets. We demonstrate that with only 600M parameters, SmolRGPT achieves competitive results on challenging warehouse spatial reasoning benchmarks, matching or exceeding the performance of much larger alternatives. These findings highlight the potential for efficient, deployable multimodal intelligence in real-world settings without sacrificing core spatial reasoning capabilities. The code of the experimentation will be available at: \href{https://github.com/abtraore/SmolRGPT}{https://github.com/abtraore/SmolRGPT} 
\end{abstract}
    
\section{Introduction}

Vision-language models (VLMs) have become the foundation for many multimodal reasoning tasks, ranging from image captioning to visual question answering and embodied AI. Recent advances have seen large, generalist models achieving impressive results by combining powerful language models with sophisticated visual encoders~\cite{alayrac2022flamingo, li2023blip2, Liu2023}. However, the substantial computational resources and high memory requirements of these models present significant challenges for practical deployment in resource-constrained environments, such as warehouses, robotics, or industrial systems.
Crucially, such real-world settings impose distinct requirements: models must not only conduct efficient reasoning under strict hardware and memory limits, but also exhibit robust spatial understanding and reasoning about where objects are, how they are arranged, and what geometric relationships they exhibit. Yet, despite progress in scaling and generality, recent studies highlight that spatial reasoning remains a key weakness for most VLMs~\cite{Kamath2023, wang2024spatial}. Several benchmarks now reveal that these models trail far behind human-level or even text-only model performance on basic spatial queries.
To bridge this gap, benchmark datasets~\cite{Chen2024b, Kamath2023, wang2024spatial} and emerging architectures~\cite{Cheng2024, Guo2024a} have targeted explicit spatial understanding via 3D scene graphs, region-aware prompts, or depth-augmented input. However, most approaches demonstrating strong reasoning capabilities still rely on large model sizes, often exceeding 2B or even 8B parameters, making them impractical for real-time execution at the edge. Simultaneously, there is growing interest in the development of memory and compute-efficient architectures for vision-language tasks~\cite{Marafioti2025, Yao2024}, with models as small as a few hundred million parameters now achieving promising performance on various benchmarks. Notably, few works have successfully combined both the architectural efficiency needed for practical deployment and the complex spatial capabilities demanded by logistics and warehouse robotics applications.

In this work, we introduce SmolRGPT, a region-aware vision-language model that explicitly incorporates spatial relationships using both RGB and depth cues, while maintaining a compact parameter footprint suitable for constrained settings. Through an efficient training curriculum and architecture, we demonstrate that region-level spatial reasoning can be achieved without resorting to multibillion-parameter models. Our results on challenging warehouse-focused tasks represent a step toward deployable spatial intelligence for real-world applications.
\section{Related Work}

Vision-language models have emerged as a powerful paradigm for multimodal understanding, with recent advances focusing on both scale and efficiency. Large-scale VLMs such as Flamingo~\cite{alayrac2022flamingo}, BLIP-2~\cite{li2023blip2}, and LLaVA~\cite{Liu2023} have demonstrated impressive capabilities by leveraging billions of parameters. However, the computational demands of these models limit their deployment in resource-constrained environments such as warehouses and industrial settings.

To address these challenges, recent work has increasingly emphasized efficient VLMs suitable for edge deployment. SmolVLM~\cite{Marafioti2025} shows that models with as few as 256M parameters can achieve competitive performance while using less than 1GB of GPU memory, outperforming models 300$\times$ larger through pixel shuffle tokenization and architectural optimizations. MiniCPM-V 2.6~\cite{Yao2024} attains GPT-4V-level performance with 8B parameters, utilizing 75\% fewer visual tokens via perceiver resampler architectures. Moondream2~\cite{vik_2024} and Florence-2~\cite{Xiao2023} further show that models under 2B parameters can achieve strong visual understanding while remaining deployable on mobile devices. These efficiency-focused approaches often target model sizes in the sub-billion or low billion parameter range to balance performance with deployment constraints in real-world environments.

Despite significant progress in visual understanding, spatial reasoning remains a fundamental challenge for VLMs. SpatialVLM~\cite{Chen2024b} addresses this limitation by introducing an Internet-scale dataset with 2 billion spatial VQA examples, demonstrating that VLMs can learn 3D spatial relationships from 2D images when provided with sufficient training data. However, this approach relies on large-scale models and extensive computational resources.

Recent evaluations have also revealed notable limitations in the spatial reasoning capabilities of VLMs. Kamath et al.~\cite{Kamath2023} show that VLMs achieve only 56\% accuracy on basic spatial relations (up/down, left/right), compared to 99\% human performance. Wang et al.~\cite{wang2024spatial} demonstrate that VLMs often underperform text-only LLMs on spatial reasoning tasks. The SpatialEval~\cite{wang2024spatial} and What's~Up~\cite{Kamath2023} benchmarks provide comprehensive evaluations highlighting these fundamental gaps.

Among efforts to enhance spatial reasoning, SpatialRGPT~\cite{Cheng2024} proposes a plugin-based architecture for flexible depth integration and regional representation learning from 3D scene graphs, achieving improvements on spatial reasoning benchmarks with an 8B parameter model. RegionGPT~\cite{Guo2024a} enhances spatial awareness in visual encoders via task-guided instruction prompts and automated region caption generation. GLaMM~\cite{Rasheed2023} introduces pixel-level grounded conversations with over 810M region annotations, while LISA~\cite{Lai2023} proposes reasoning segmentation that outputs masks based on complex queries requiring world knowledge.

Region-level understanding has also been extended to 3D environments: VLM-Grounder~\cite{Xu2024} achieves zero-shot 3D visual grounding using only 2D images via dynamic image stitching and multi-view ensemble projection; VLM-3R~\cite{Fan2025} demonstrates robust 3D spatial reasoning from monocular RGB video without external depth sensors; and PointLLM~\cite{Xu2023b} enables spatial understanding by directly processing colored 3D point clouds.

Although these approaches demonstrate sophisticated spatial grounding capabilities, they typically require large model architectures or complex, multi-stage pipelines. The AI City Challenge 2025 Track 3 \cite{Tang25AICity25} specifically highlights the need for practical spatial intelligence in warehouse environments, where deployment constraints and real-time requirements require efficient architectures. This ongoing research direction seeks to demonstrate that small, well-designed models can achieve competitive performance on complex spatial reasoning tasks when combined with appropriate training strategies and architectural innovations.

\section{Methodology}
\label{sec:methods}

SmolRGPT builds upon SmolVLM by integrating region-level visual language modeling techniques from SpatialRGPT. We began our implementation with NanoVLM~\cite{wiedmann2025nanovlm}, which provides a basic VLM framework from scratch. Our approach uses a pretrained visual feature extractor, siglip2-base-patch16-256 ~\cite{Tschannen2025}, to process both RGB and depth images. This extractor outputs fixed-size feature vectors of dimension 256: $I_v$ for RGB and $D_v$ for depth.

First, the RGB feature vector $I_v$ passes through an RGB Connector, which consists of pixel shuffling followed by a linear layer, mapping it into the language model's embedding space.
Next, the projected features are upsampled to match the mask resolution in the subsequent stage using transpose convolutions within dedicated refiner modules, whose design is similar to that of \cite{Guo2024a}.
This process is performed independently for the RGB features ($I_v$) and the depth features ($D_v$), using modality-specific refiners to produce the refined features $I_r$ and $D_r$, respectively.
After refinement, both $I_r$ and $D_r$ undergo mask pooling with region masks to extract features for specific regions of interest.
These region-level features are then inserted into the embedding sequence of the pretrained language model, SmolLM2-360M~\cite{Allal2025}.
By maintaining separate projection and refinement pathways for RGB and depth inputs, this design follows the approach of SpatialGPT, ensuring the representations for each modality remain distinct and avoiding confusion between RGB and depth features.

\begin{figure*}
  \centering
    \includegraphics[width=1\linewidth]{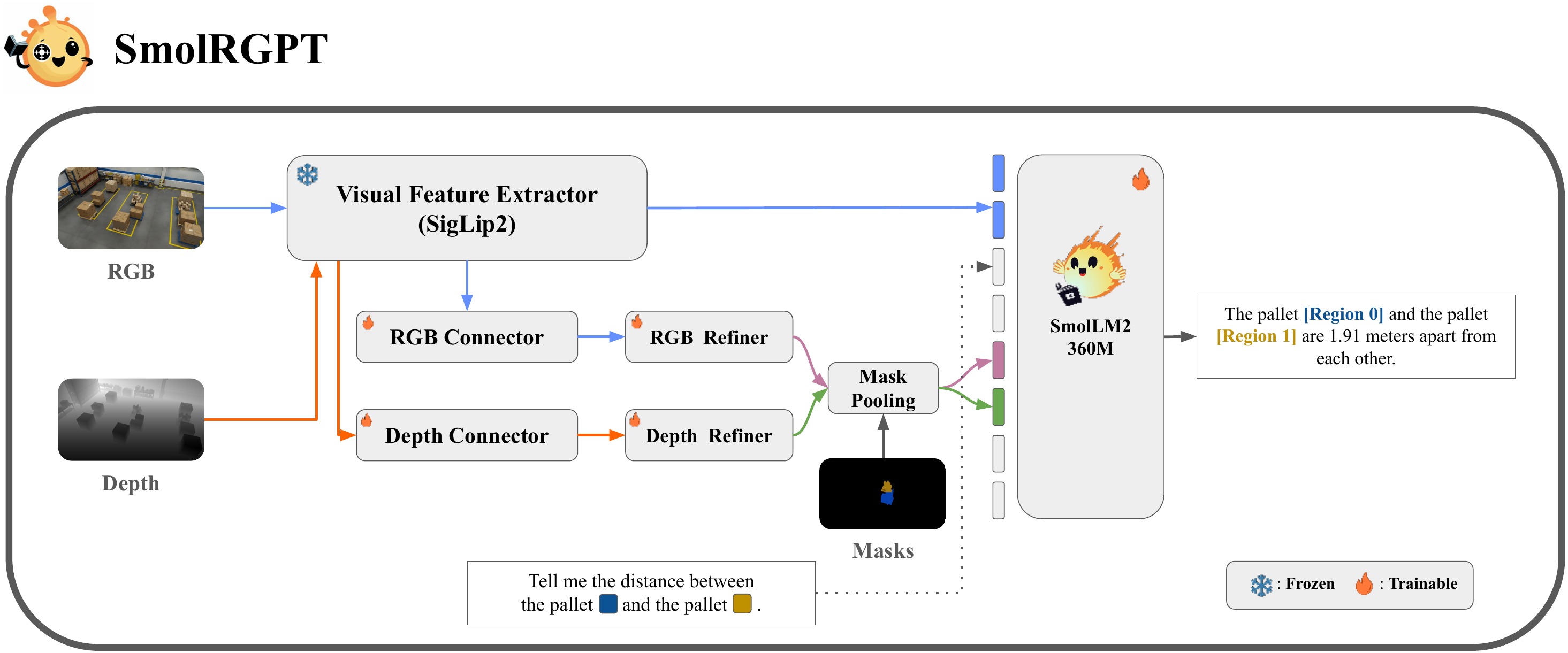}
    \caption{SmolRGPT architecture overview. The model processes RGB and depth images through a shared Visual Feature Extractor (SigLip2) followed by modality-specific pathways. RGB features pass through an RGB Connector and RGB Refiner (shown in blue), while depth features are processed by a Depth Connector and Depth Refiner (shown in orange). Both refined features undergo Mask Pooling with region masks to extract spatial representations. These pooled features, along with the visual features, are integrated with the SmolLM2 360M language model to generate spatial reasoning responses. The example shows the model answering a distance query between warehouse regions. The frozen components (Visual Feature Extractor) are indicated by the snowflake symbol, while trainable components are marked with fire symbols.}
    \label{fig:smol}
\end{figure*}

It's important to underline that SmolRGPT's connectors differ from the ones used in SpatialRGPT: it employs pixel shuffling, which rearranges spatial features into the channel dimension.
This reduces spatial resolution, but increases channel depth, providing denser representations.
This overall design enables SmolRGPT to effectively combine both global and region-specific visual information, improving its region-level language modeling abilities.
Figure~\ref{fig:smol} shows the overall architecture of SmolRGPT.

\subsection{Tokenization and Prompt Format}
\begin{figure*}
  \centering
    \includegraphics[width=1\linewidth]{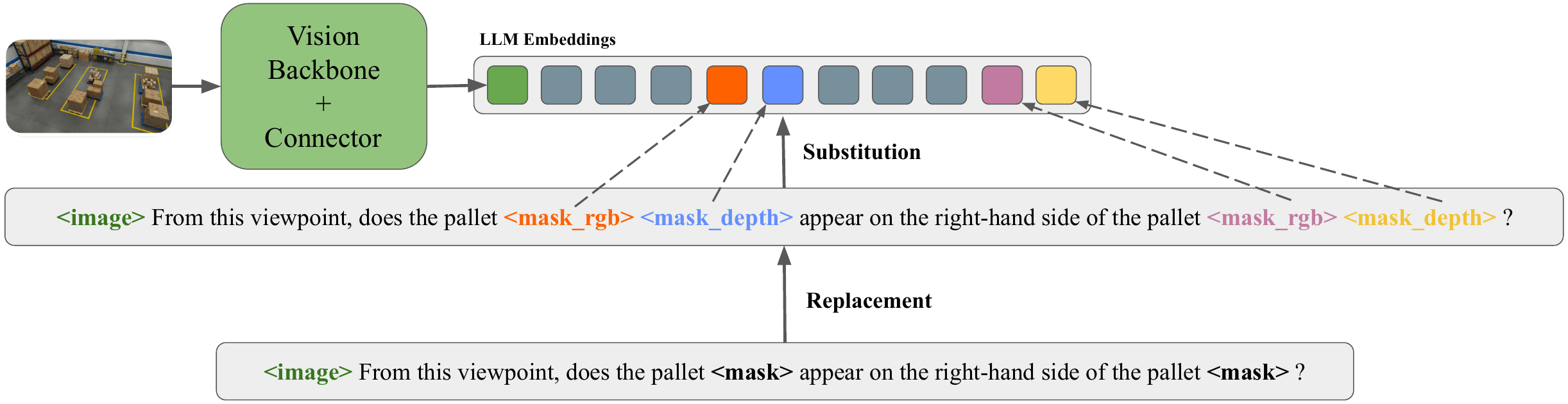}
    \caption{Token substitution and replacement mechanism in SmolRGPT. The vision backbone and projector convert the warehouse image into LLM embeddings, producing a sequence of visual tokens. During processing, special tokens in the input prompt (\texttt{<mask\_rgb>} and \texttt{<mask\_depth>}) are identified and replaced with their corresponding mask-pooled regional embeddings. The example illustrates how the model handles a spatial query containing multiple mask tokens, where each token is substituted with the appropriate region-specific features extracted through the mask pooling mechanism. This approach enables precise spatial referencing within the warehouse scene while maintaining the natural language interface of the language model.}
    \label{fig:embeddings}
\end{figure*}

Multi-turn conversation formats are used following~\cite{Liu2023, Guo2024a, Cheng2024}. In all datasets, the \texttt{<image>} token in the input sequence is replaced by the global image features, following established protocols. For region-level datasets such as the Warehouse dataset and OSD (Open Spatial Dataset)~\cite{Cheng2024}, we introduce additional specialized tokens: each \texttt{<mask>} placeholder in the data is replaced with \texttt{<mask\_rgb>} and \texttt{<mask\_depth>}. These new tokens act as placeholders that are subsequently substituted with their respective region-specific RGB ($I_r$) and depth ($D_r$) embeddings from the refiners.
The resulting visual embeddings, interpreted as special tokens, are interleaved with textual tokens and together form the input sequence for the LLM. Figure~\ref{fig:embeddings} illustrates this process for visual region-level datasets.
In the case of global-image datasets such as LLaVA-CC3M~\cite{Liu2023}, only the \texttt{<image>} token is present; it is replaced with the corresponding global image sequence embeddings, as no region-level segmentation is provided. 

\subsection{Training Scheme}
\label{sec:experiments_training}

SmolRGPT was trained in three sequential stages: (1) RGB connector alignment, (2) depth connector and refiner warmup, and (3) supervised finetuning. In the first stage, all model components were frozen except for the RGB connector, which was trained on the LLaVA-CC3M ~\cite{Liu2023} dataset, a chat-formatted, filtered subset of CC3M containing 595k image-text pairs for image captioning. This stage develops the understanding of the global scene of the model, with a relatively high learning rate of $10^{-4}$ selected to facilitate effective learning in this newly initialized module.
In the second stage, the RGB connector was frozen, while the depth connector and both the RGB and depth refiners were unfrozen. Training took place on the first one million examples from the Open Spatial Dataset (OSD)~\cite{Cheng2024}, using the same learning rate of $10^{-4}$.
The goal in that stage was to initialize these components in preparation for final finetuning.
Output quality at this stage was intentionally suboptimal, as it served primarily as a warmup step.
We limited training to one million OSD samples for efficiency, since the dataset closely matches the data used in the next stage and a full pass over all 8.7 million OSD samples would be prohibitively time-consuming on our system that is only equipped with two RTX 5090 GPUs.
In the third and final stage, all components except the image backbone were unfrozen and jointly trained on the warehouse dataset \cite{Tang25AICity25}, which contains over 500k training samples. Here, we used a lower learning rate of $5 \times 10^{-5}$ to prevent catastrophic forgetting and maintain training stability.

\subsection{Evaluation Scheme}

As the outputs of the supervised finetuned models for region-level modeling are sentences, it is challenging to evaluate them both quantitatively and qualitatively.
For example, given a question such as: ``From this viewpoint, does the pallet \texttt{<mask>} appear on the right-hand side of the pallet \texttt{<mask>}?'', the expected answer may be ``left'' or ``right'', but the model might respond with a sentence like: ``From this viewpoint, the pallet [Region 0] is on the left of the pallet [Region 1].''
It is important to note that such outputs are not inherently wrong; rather, they are not normalized in a way that allows the computation of evaluation metrics.
To properly evaluate model performance, it is therefore essential to normalize these outputs into a standard form suitable for quantitative assessment.

In the context of the warehouse dataset, we define four question types: left-right, count, distance, and multiple-choice grounding.
To facilitate answer normalization, we train two Longformer models~\cite{Beltagy2020} to classify the type of question and the type of answer.
Based on this information, we use Qwen2.5-14B~\cite{Qwen2024} running on Ollama~\cite{githubGitHubOllamaollama}, and apply prompts tailored to each question and answer type, enabling us to extract normalized answers.
These normalized answers can then be used to compute various metrics such as accuracy or root mean square error.

It is important to emphasize that the Longformer models and Qwen are used solely for evaluation purposes, and are not part of the model's inference pipeline.

\begin{figure*}
  \centering
    \includegraphics[width=1\linewidth]{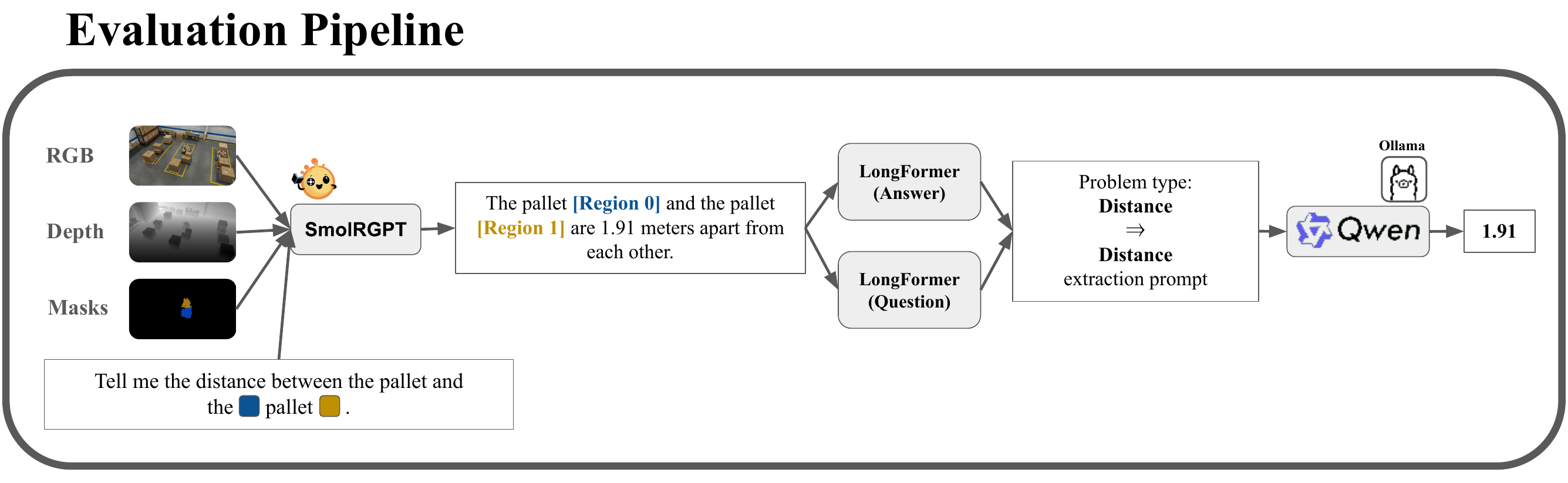}
    \caption{SmolRGPT evaluation pipeline for spatial reasoning in warehouse environments. The system processes RGB images, depth maps, and region masks through the SmolRGPT model to generate natural language responses to spatial queries. The generated responses undergo a two-stage extraction process: first, separate classifiers determine the answer type from both the question (Longformer Question) and answer (Longformer Answer), then Qwen2.5-14B extracts the normalized answer (e.g., "1.91") based on the identified problem type (Distance). This pipeline ensures robust answer extraction across different spatial reasoning categories including distance estimation, counting, spatial relations, and multiple-choice questions. The example demonstrates distance extraction from a response about pallets in a warehouse scene.}
    \label{fig:inference}
\end{figure*}

\section{Datasets}
\label{sec:datasets}

Our approach to developing SmolRGPT leverages a carefully curated progression of datasets, each serving a specific purpose in building robust spatial reasoning capabilities for warehouse environments. We employ a three-stage training curriculum that mirrors the complexity gradient from general vision-language alignment to specialized spatial understanding, culminating in warehouse-specific task performance. This section describes the three primary datasets utilized in our training pipeline: LLaVA-CC3M-595K~\cite{Liu2023} for initial vision-language alignment, the Open Spatial Dataset (OSD)~\cite{Cheng2024} for general spatial reasoning capabilities, and the PhysicalAI-Spatial-Intelligence-Warehouse dataset \cite{Tang25AICity25} for task-specific fine-tuning. Each dataset contributes unique characteristics that collectively enable our 600M parameter model to achieve competitive performance on complex warehouse spatial reasoning tasks while maintaining computational efficiency.

\subsection{LLaVA-CC3M-595K Dataset}
For our experiments, we utilize the LLaVA Visual Instruct CC3M Pretrain 595K dataset~\cite{Liu2023}, a carefully curated subset of 595,000 image-text pairs derived from the original Conceptual Captions 3M (CC-3M) dataset~\cite{sharma2018conceptual}. This subset, representing approximately 20\% of the original corpus, was filtered to achieve a more balanced concept coverage distribution, addressing potential biases where certain visual concepts may be overrepresented in CC-3M. Each sample contains the original image, its corresponding human-generated caption, and an additional BLIP~\cite{Li2022} synthetic caption included for reference and comparison purposes. The dataset is specifically constructed for the pretraining stage of visual instruction tuning, where it facilitates feature alignment between vision and language modalities. During this stage, only the projection matrix connecting the frozen vision encoder and language model is trained, enabling efficient feature alignment using a relatively compact dataset compared to other multimodal pretraining approaches. The balanced concept distribution ensures the model encounters diverse visual scenarios during pretraining, while the inclusion of both original and synthetic captions provides flexibility in training strategies and enables investigation of caption quality effects on alignment performance.

\subsection{Open Spatial Dataset}
To enhance our model's spatial reasoning capabilities, we leverage insights from the Open Spatial Dataset (OSD) from SpatialRGPT~\cite{Cheng2024}, which provides complementary large-scale 3D spatial annotations. The OSD contains 8.7 million spatial concepts grounded in 5 million unique regions from 1 million images, offering extensive pretraining data for spatial understanding. While OSD focuses on general spatial reasoning across diverse scenes, its automated 3D scene graph construction pipeline and metric-space grounding provide valuable foundations for warehouse-specific spatial intelligence.
The integration of OSD's spatial representation learning benefits our approach in several key ways. The dataset's comprehensive coverage of distance relationships, directional concepts, and size comparisons aligns well with the warehouse spatial reasoning requirements. Its region-aware processing capabilities, which enable precise spatial localization and relationship queries between user-specified regions, directly transfer to warehouse scenarios where accurate spatial referencing is critical. Furthermore, the metric-scale spatial understanding developed through OSD pretraining provides a strong foundation for the precise dimensional and distance estimations required in industrial environments.

\subsection{Warehouse Dataset}
For the AI City Challenge 2025 Track 3, we utilize the PhysicalAI-Spatial-Intelligence-Warehouse dataset, which serves as a pioneering benchmark for evaluating spatial reasoning in industrial environments \cite{Tang25AICity25}.
This dataset addresses a critical gap in AI systems by focusing on the ability to understand detailed spatial relationships within warehouse-scale operational settings, moving beyond tasks limited to synthetic or domestic scenes. It comprises real-world warehouse scenes with comprehensive 3D object layouts, accurate dimensional information, and intricate spatial relationships, alongside natural language question-answer pairs crafted to assess spatial reasoning capabilities.

The dataset’s questions are organized into four primary categories to test different aspects of spatial intelligence. Distance queries assess the understanding of metric spatial relationships within the warehouse. Count questions evaluate the ability to enumerate objects contained within specific areas. Multiple-choice grounding tasks challenge models to make relative spatial comparisons and identify regions with respect to given references. Spatial relation queries test the comprehension of topological and directional relationships between objects. This diverse set of questions types ensures a thorough evaluation of both quantitative and qualitative spatial reasoning abilities.

To support consistent assessment, the dataset offers standardized training, validation, and test splits, with normalized answer formats. Each instance includes objects indexed by region within the warehouse, allowing for precise spatial referencing via a unified coordinate system. The normalization protocol accommodates common variations in response formats, supporting a range of numeric and word representations while maintaining consistency across different types of queries. This structure enables robust benchmarking of various model architectures, ranging from modular agent-based systems to end-to-end approaches combining 3D vision and language understanding.

\section{Experiments and Discussion}
\label{sec:experiments}


In this section, we present a comprehensive evaluation of SmolRGPT and compare its performance with state-of-the-art large vision-language models on domain-specific spatial reasoning tasks drawn exclusively from the AI City Challenge 2025 Warehouse Track \cite{Tang25AICity25}.
We analyze the effectiveness of our compact 600M-parameter architecture in terms of accuracy, parameter efficiency, and generalizability within this warehouse context.
Performance is reported for the specialized warehouse scenarios of the challenge, enabling a clear understanding of the model’s capabilities in a realistic industrial setting. Our findings demonstrate that careful architectural and training choices can enable small models to achieve competitive or even superior results in complex spatial tasks traditionally dominated by much larger models.

\subsection{Training Details}

\paragraph{SmolRGPT.}
We employed the three-stage training scheme described in Section~\ref{sec:experiments_training}. In the first stage, we trained on the LLaVA-CC3M dataset for 48{,}379 steps (equivalent to 20 epochs), using a per-device batch size of 30 and 4 steps of gradient accumulation on 2$\times$5090 GPUs, resulting in an effective batch size of 240.
In the second stage, we switched to our subset of the OSD dataset, running for 4{,}787 steps (1 epoch) with a slightly reduced per-device batch size of 28 (effective batch size 224); this reduction was necessary due to the increased memory requirements from unfreezing both refiners and the depth connector.
The final stage involved training on the warehouse dataset, where we initially planned for 20 epochs but observed that model learning plateaued after epoch 4 (18{,}000 steps), and the validation loss increased up to epoch 9 (40{,}339 steps); accordingly, the batch size was further reduced to 14 per device (effective batch size 112) to accommodate memory constraints. Throughout all stages, we used the AdamW optimizer with stage-specific learning rates (see Section~\ref{sec:experiments_training}), a weight decay of 0.01, a cosine learning rate scheduler, linear warmup over 3\% of total training steps for each stage, and enabled automatic mixed precision in PyTorch to optimize VRAM usage.
\paragraph{Longformers.}
For our inference pipeline, we trained two Longformer models on OSD training and validation splits: one for question classification and one for answer classification. Both models were trained for 250 steps with a batch size of 128 on the dual 5090 GPU setup, using the AdamW optimizer (learning rate $2\times10^{-5}$, weight decay 0.01), mixed precision, and gradient checkpointing to further reduce VRAM usage.


\subsection{Warehouse}

SmolRGPT achieved 3rd place in the AI City Challenge 2025 Track 3 with a final S1 score of 90.68 as shown in table \ref{tab:leaderboard} on the full test set, demonstrating that efficient 600M parameter models can compete effectively against substantially larger architectures. The model maintained exceptional performance on spatial relationship understanding, achieving 99.80\% accuracy on left-right directional tasks, confirming that compact architectures can capture precise spatial semantics when properly designed. 
Counting tasks showed strong performance with 92.76\% accuracy and an RMSE of 0.0750, validating our approach of integrating depth information through dedicated refiners for improved object separation in cluttered warehouse environments. Multiple-choice questions, which require holistic scene understanding and reasoning, achieved 88.02\% accuracy, demonstrating robust comprehension of complex spatial queries. 
Distance estimation remained the most challenging task with 82.13\% accuracy and an RMSE of 0.4740. While this represents our lowest individual metric, it significantly outperforms what would be expected from a 600M parameter model without proper depth integration, validating our architectural decision to include separate RGB and depth refiners.
These results on the full test set confirm that SmolRGPT successfully balances efficiency with performance. While spatial reasoning models like SpatialRGPT require 8B parameters for general spatial understanding tasks, our model achieved competitive results on warehouse-specific spatial reasoning with 13× fewer parameters while remaining deployable on consumer hardware. This 3rd place among all participating teams validates our hypothesis that carefully designed efficient architectures, combined with appropriate training strategies, can achieve strong performance on complex spatial reasoning tasks without requiring massive computational resources.
\begin{table}[t]
  \centering
  \begin{tabular}{@{}l l c@{}}
    \toprule
    Rank & Team Name & Score \\
    \midrule
    1 & UWIPL\_ETRI & 96.0789 \\
    2 & HCMUT.VNU & 91.9735 \\
    \textbf{}{3} & \textbf{Embia} & \textbf{90.6772} \\
    4 & MIZSU & 73.0606 \\
    5 & HCMUS\_HTH & 66.8861 \\
    6 & MealsRetrieval & 56.6352 \\
    7 & BKU22 & 50.3662 \\
    8 & Smart Lab & 31.9245 \\
    9 & AICV & 28.2993 \\
    \bottomrule
  \end{tabular}
  \caption{Leaderboard results.}
  \label{tab:leaderboard}
\end{table}

\subsection{General Spatial Reasoning}
To further test our methodology on a wider range of scenarios, we decided to train our best model on OSD for 1 epoch with a mix of the Warehouse Dataset to avoid catastrophic forgetting and test it on SpatialRGPT-Bench~\cite{Cheng2024}. The results are available in tables \ref{tab:result_osd_qualitative} and \ref{tab:result_osd_quantitative}.

\subsubsection{Qualitative Spatial Reasoning Results}

Table \ref{tab:result_osd_qualitative} presents the performance of various models on qualitative spatial relation tasks. A critical observation is the remarkable efficiency of our approach: SmolRGPT, with only 600M parameters, achieves an average accuracy of 65.6\% across all spatial relation categories. This significantly outperforms much larger models including GPT-4 (~1.76T parameters, 57.8\%), GPT-4V (~1.76T parameters, 58.1\%), and LLaVA-v1.6-34B (34B parameters, 43.9\%). The parameter efficiency ratio is striking - SmolRGPT achieves better performance with approximately 2,933× fewer parameters than GPT-4 and 57× fewer parameters than LLaVA-v1.6-34B.
SmolRGPT demonstrates particularly strong performance on the Behind/Front relation (79.0\%) and Tall/Short relation (74.1\%), suggesting robust understanding of depth and height relationships in images despite its compact size. This efficiency makes SmolRGPT particularly suitable for deployment in resource-constrained environments such as mobile devices, edge computing platforms, and embedded systems where computational resources and energy consumption are critical constraints.
While SpatialRGPT and SpatialRGPT-Depth achieve superior performance (89.8\% and 91.7\% average accuracy respectively), it is important to note that these models are specifically designed and extensively trained for spatial reasoning tasks with significantly larger architectures. In contrast, SmolRGPT represents a more general and efficient approach that balances spatial understanding with computational efficiency. The performance gap between SmolRGPT and models enhanced with Set-of-Mark (SoM) prompting (GPT-4V+SoM: 54.3\%, LLaVA-v1.6-34B+SoM: 42.3\%) demonstrates that our training approach effectively incorporates spatial understanding directly into the model rather than relying on external visual prompting techniques.
Kosmos-2, despite having approximately 1.3B parameters (more than twice the size of SmolRGPT), achieves only 17.0\% average accuracy, highlighting that model size alone does not guarantee spatial reasoning capabilities. This further emphasizes the effectiveness of our training methodology in achieving strong performance with minimal parameters.

\subsubsection{Qualitative Spatial Reasoning Results}
Table \ref{tab:result_osd_quantitative} evaluates the models' ability to estimate quantitative spatial properties. Here, we observe more varied performance across different metrics. SmolRGPT achieves competitive results on Direct Distance (35.8\%) and Direction (35.5\%) tasks, matching or exceeding the performance of much larger models. For instance, SmolRGPT outperforms GPT-4V (29.7\%) on Direct Distance estimation despite having ~2,933× fewer parameters. However, the model shows lower performance on Width (18.05\%) and Height (20.30\%) estimation tasks, suggesting areas for future improvement.
The quantitative results reveal interesting patterns: while larger models like GPT-4V excel at estimating object dimensions (Width: 51.1\%, Height: 68.4\%), they struggle with distance-based measurements relative to their size. SmolRGPT's balanced performance across distance and direction tasks, despite its compact 600M parameter architecture, indicates that our training methodology successfully captures essential spatial relationships without requiring the computational overhead of billion or trillion-parameter models.
These results collectively demonstrate that SmolRGPT provides an unprecedented balance between model efficiency and spatial reasoning capability. With only 600M parameters, it achieves performance comparable to or better than models that are orders of magnitude larger, making it particularly suitable for deployment in resource-constrained environments where spatial understanding is required but computational resources, memory, and energy consumption are limited. This efficiency breakthrough opens new possibilities for deploying spatial AI capabilities on edge devices, mobile platforms, and embedded systems.

\begin{table*}[t]
\centering
\footnotesize
\begin{tabular}{@{}lccccccccc@{}}
\toprule
\textbf{Method} & \textbf{Parameters} & \textbf{Below/Above} & \textbf{Left/Right} & \textbf{Big/Small} & \textbf{Tall/Short} & \textbf{Wide/Thin} & \textbf{Behind/Front} & \textbf{Avg.} \\
\midrule
GPT-4 \cite{OpenAI_GPT4_2023} & 1.76T & 64.1 & 42.8 & 42.8 & 61.6 & 61.6 & 49.0 & 57.8 \\
GPT-4V \cite{Yang2023a} & 1.76T & 63.3 & 46.6 & 64.1 & 60.7 & 68.2 & 45.4 & 58.1 \\
LLaVA-v1.6-34B \cite{liu2024llavanext} & 34B & 44.1 & 45.7 & 36.7 & 53.5 & 37.5 & 45.4 & 43.9 \\
GPT-4V+SoM \cite{Yang2023a} & 1.76T & 75.0 & 55.2 & 42.4 & 54.4 & 49.0 & 47.2 & 54.3 \\
LLaVA-v1.6-34B+SoM \cite{liu2024llavanext} & 34B & 44.1 & 40.0 & 33.9 & 47.3 & 41.3 & 46.3 & 42.3 \\
Kosmos-2 \cite{Peng2023} & 1.3B & 28.3 & 15.2 & 4.71 & 26.7 & 12.5 & 12.7 & 17.0 \\
RegionVILA \cite{Guo2024a} & 7B$^\dagger$ & 30.8 & 47.6 & 35.8 & 44.6 & 35.5 & 49.0 & 40.4 \\
\rowcolor{yellow!50}
\textbf{SmolRGPT} & \textbf{600M} & 71.6 & 49.5 & 67.9 & 74.1 & 51.9 & 79.0 & 65.6 \\
SpatialRGPT \cite{Cheng2024} & 7B$^\dagger$ & 99.1 & 99.0 & 79.2 & 89.2 & 83.6 & 87.2 & 89.8 \\
SpatialRGPT-Depth \cite{Cheng2024} & 7B$^\dagger$ & 99.1 & 99.0 & 80.1 & 91.9 & 87.5 & 91.8 & 91.7 \\
\bottomrule
\end{tabular}
\caption{Accuracy (\%) on qualitative spatial relation tasks across different models. Each value indicates the percentage of correct predictions for each spatial relation category (higher is better). SmolRGPT (highlighted) is our method with only 600M parameters. The last column (Avg.) reports the average accuracy across all qualitative categories. $^\dagger$Estimated based on LLaMA-2 7B backbone.}
\label{tab:result_osd_qualitative}
\end{table*}

\begin{table*}[ht]
\centering
\footnotesize
\begin{tabular}{lccccccc}
\toprule
\textbf{Method} & \textbf{Parameters} & \textbf{Direct Distance} & \textbf{Horizontal Distance} & \textbf{Vertical Distance} & \textbf{Width} & \textbf{Height} & \textbf{Direction} \\
\midrule
GPT-4 \cite{OpenAI_GPT4_2023} & 1.76T & 21.6 & 11.5 & 33.0 & 52.3 & 48.1 & 34.6 \\
GPT-4V \cite{Yang2023a} & 1.76T & 29.7 & 25.4 & 33.0 & 51.1 & 68.4 & 43.9 \\
LLaVA-v1.6-34B \cite{liu2024llavanext} & 34B & 24.3 & 24.5 & 30.4 & 30.8 & 42.8 & 33.6 \\
GPT-4V+SoM \cite{Yang2023a} & 1.76T & 25.7 & 22.1 & 33.9 & 45.8 & 62.4 & 54.2 \\
LLaVA-v1.6-34B+SoM \cite{liu2024llavanext} & 34B & 12.8 & 20.4 & 11.3 & 9.02 & 7.52 & 11.3 \\
Kosmos-2 \cite{Peng2023} & 1.3B & 4.05 & 4.91 & 18.9 & 3.01 & 3.10 & 3.82 \\
RegionVILA \cite{Guo2024a} & 7B$^\dagger$ & 22.3 & 24.6 & 17.9 & 36.8 & 49.6 & 35.5 \\
\rowcolor{yellow!50}
\textbf{SmolRGPT} & \textbf{600M} & 35.8 & 18.3 & 33.9 & 18.05 & 20.3 & 35.5 \\
SpatialRGPT \cite{Cheng2024} & 7B$^\dagger$ & 35.1 & 59.0 & 53.8 & 51.9 & 54.9 & 95.3 \\
SpatialRGPT-Depth \cite{Cheng2024} & 7B$^\dagger$ & 41.2 & 65.6 & 51.9 & 49.6 & 57.9 & 95.3 \\
\bottomrule
\end{tabular}
\caption{SpatialRGPT-Bench Quantitative Results: Numbers represent success rates within $\pm$25\% of ground-truth in percentage. Quantitative performance on spatial reasoning tasks with model parameter counts. SmolRGPT (highlighted) achieves competitive performance with only 600M parameters. $^\dagger$Estimated based on LLaMA-2 7B backbone.}
\label{tab:result_osd_quantitative}
\end{table*}

\section{Conclusion}

SmolRGPT demonstrates that strong region-level spatial reasoning can be achieved without the scale of multi billion parameter vision language models.  
The key takeaways are:

\begin{itemize}
    \item \textbf{Parameter–performance trade-off.}  
          With only 600\,M parameters, SmolRGPT ranks 3\textsuperscript{rd} on the AI City Challenge 2025 Warehouse Track and matches or exceeds GPT-4(V) and LLaVA-34B on several qualitative spatial relations.
    \item \textbf{Architectural contributions.}  
          Separate RGB and depth connectors and refiners, a pixel shuffle projection, and explicit region token insertion give the model precise left right, front behind and distance reasoning while keeping sequence length modest.
    \item \textbf{Curriculum design.}  
          The three stage schedule (global alignment on CC3M, spatial warm up on OSD, task specific fine tuning on the warehouse data) builds spatial competence progressively and avoids catastrophic forgetting.
    \item \textbf{Current limitations.}  
          Absolute size estimation (width and height) still trails state of the art, the answer extraction pipeline is external to the model, and robustness to domain shift or changing layouts remains untested.
\end{itemize}

Careful architectural choices and a progressive training curriculum allows SmolRGPT to narrow much of the gap between compact models and very large VLMs while keeping the parameter count modest.  
These results suggest that efficient and deployable multimodal spatial reasoning is feasible without the heavy computational overhead that has often been viewed as necessary.

{
    \small
    \bibliographystyle{ieeenat_fullname}
    \bibliography{main}
}

\end{document}